\documentclass[11pt,letterpaper]{article}
\usepackage{emnlp2017}
\usepackage{times}
\usepackage{latexsym}

\usepackage{graphicx,tabulary,multicol,multirow,booktabs,array,xcolor,wrapfig,cuted}

\usepackage[lofdepth,lotdepth]{subfig}

\usepackage[draft,inline,nomargin,index]{fixme}
\fxsetup{theme=color,mode=multiuser}
\FXRegisterAuthor{es}{esn}{\color{red}ES}
\FXRegisterAuthor{ec}{ecn}{\color{blue}EC}
\FXRegisterAuthor{al}{aln}{\color{green}AL}

\usepackage{linguex}

\usepackage{makeidx}
\makeindex

\emnlpfinalcopy

\def\emnlppaperid{84}

\title{Measuring Thematic Fit with Distributional Feature Overlap}

\author{Enrico Santus\textsuperscript{1}, Emmanuele Chersoni\textsuperscript{2}, Alessandro Lenci\textsuperscript{3}\and Philippe Blache\textsuperscript{2} \\
  {\tt enrico\_santus@sutd.edu.sg} \\
  {\tt emmanuelechersoni@gmail.com} \\
  {\tt alessandro.lenci@unipi.it} \\
  {\tt philippe.blache@univ-amu.fr} \\
  \textsuperscript{1} Singapore University of Technology and Design \\
  \textsuperscript{2} Aix-Marseille University \\
  \textsuperscript{3} University of Pisa
  }

\date{}

\begin{document}

\maketitle

\begin{abstract}
  %In this paper we introduce APFit, a distributional method for estimating how well a word fits the agent (and patient) role of a specific a predicate. This task, known in the literature as thematic fit estimation, is gaining a lot of attention in the last years for its linguistic and cognitive relevance (e.g. for the characterization of events). However, so far, the number of studies that have dealt with it in a completely unsupervised way can be counted on one hand. 
%\esnote[margin,noinline]{Let's not forget to update the template to the ACL one}
In this paper, we introduce a new distributional method for modeling predicate-argument thematic fit judgments.
%, working purely on syntactic information.
%, inspired by several cognitive and psycholinguistic findings.\esnote[margin,noinline]{How about removing cognitve findings from here and highlighting instead we don't use semantic information but only dep one?} 
We use a syntax-based DSM to build a prototypical representation of verb-specific roles: for every verb, we extract the most salient \textit{second order contexts} for each of its roles (i.e. the most salient dimensions of \textit{typical role fillers}), and then we compute thematic fit as a \textit{weighted overlap} between the top features of candidate fillers and role prototypes.
Our experiments show that our method consistently outperforms a baseline re-implementing a state-of-the-art system, and achieves better or comparable results to those reported in the literature for the other unsupervised systems. Moreover, it provides an explicit representation of the features characterizing verb-specific semantic roles.
%\esnote[margin,noinline]{Only achieving competitive results might not be a sufficient contribution for a long paper. We need to claim something more.}
%while keeping cognitive plausibility.}
%AL: bisogna andarci piano con gli statements di cognitive plausibility
% also accounting for a number of cognitive and psycholinguistic findings.
%\ecnote[margin,noinline]{Questa cosa bisogna verificarla prima di metterla nell'abstract}
\end{abstract}

\section{Introduction}

%Whenever we read or hear a predicate, we have some expectations about the possible agent and the possible patient. For example, it would not be hard to provide a plausible answer to the question "who's arresting the thieves?". This happens because we know some linguistic and conceptual information about the predicate "to arrest" (Lenci and Lebani). In a very similar way, we would be able to answer something plausible to the question "who's the policeman arresting?", even though in this case the number of options would increase (e.g. "the thieves", "the killers", "the criminals", etc.)

%\esnote[margin,noinline]{Dobbiamo ripassare tutto il testo per chiarire slot, role, etc.}
Several psycholinguistic studies in the last two decades have brought extensive evidence that humans activate a rich array of event knowledge during sentence processing: verbs (e.g. \textit{arrest}) activate expectations about their typical arguments (e.g. \textit{cop, thief}) \citep{McRae1998ModelingTI,Altmann1999247,Ferretti2001516,McRae2005,Hare2009ActivatingEK,Matsuki2011EventbasedPI}, and nouns activate other nouns typically co-occurring in the same events \citep{Kamide2003133,bicknell2010effects}.
%Expectations about a verb argument immediately influence on-line language comprehension, because experimental subjects are able to determine the plausibility %and determine the plausibility of a candidate argument noun
Subjects are able to determine the plausibility of a noun for a given argument role and quickly use this knowledge to anticipate upcoming linguistic input \citep{mcrae2009people}. This phenomenon is referred to in the literature as \textit{thematic fit}.
Thematic fit estimation
% has recently raised interest in the Natural Language Processing community (NLP), since knowledge about thematic roles has been proved to be very useful for several NLP tasks, including question answering \cite{shen:07} and machine translation \cite{liu:10,wu:11}. 
%AL: QUESTA PARTE SOPRA NON LA METTEREI. TANTO IMMAGINO  CHE LA SOTTOMETTIAMO PER L'AREA COGNITIVE MODELLING
%EN: Ale, io la rimetterei per allargare anche un po' l'audience... Si puo sempre togliere, eventualmente
has been extensively used in sentence comprehension studies on constraint-based models, mainly as a predictor variable allowing to disambiguate between possible structural analyses.\footnote{For an overview on constraint-based models, see \citet{macdonald2016constraint}.} More in general, thematic fit is considered as a key factor in a variety of studies concerned with structural ambiguity \citep{Vandekerckhove2009ARA}.

Starting from the work of \citet{Erk2010AFC}, several distributional semantic methods have been proposed to compute the extent to which nouns fulfill the requirements of verb-specific thematic roles, and their performances have been evaluated against human-generated judgments  \citep{Baroni:2010:DMG:1945043.1945049,Lenci_composingand,Sayeed2014combining,Sayeed2015AnEO,Sayeed2016ThematicFE,Greenberg2015VerbPA,Greenberg2015ImprovingUV}.
Most research on thematic fit estimation has focused on \textit{count-based} vector representations (as distinguished from \textit{prediction-based} vectors).\footnote{We adopt the terminology from \citet{baroni2014don}.} Indeed, in their comparison between high-dimensional explicit vectors and low-dimensional neural embeddings, \citet{baroni2014don} found that thematic fit estimation is the only benchmark on which prediction models are lagging behind state-of-the-art performance.
%the latter were severely underperforming in this specific task. %More recently, since the best performing distributional models for the task relied on linguistically rich dimensions (based either on dependencies or on semantic role relations), 
This is consistent with \citet{Sayeed2016ThematicFE}'s observation that ``thematic fit modeling is particularly sensitive to linguistic detail and interpretability of the vector space".

%Although a recent neural network model by Tilk et al. \shortcite{Tilk:16} has brought an improvement over state-of-the-art systems, 
The present work sets itself among the unsupervised approaches to thematic fit estimation. %\alnote[margin,noinline]{la prima frase del paragrafo non \`e chiara} 
By relying on explicit and interpretable \textit{count-based} vector representations, %keeping the explicitness and the interpretability of the features of the vector space, 
we propose a simple, cognitively-inspired, and efficient thematic fit model using information extracted from dependency-parsed corpora. %\esnote[margin,noinline]{What about semantic? E.g. Co-occurrence? Also note: if we perform better, we should remove this, because it sounds as an excuse}. 
The key features of our proposal are \textit{a}) prototypical representations of verb-specific thematic roles, based on feature weighting and filtering of \textit{second order contexts} (i.e. contexts that are salient for many of the typical fillers of a given verb-specific thematic role), and \textit{b}) a similarity measure which computes the \textit{Weighted Overlap} ($WO$) between prototypes and candidate fillers.\footnote{Code: https://github.com/esantus/Thematic\_Fit}

\section{Related Work}
\label{related_work}

\citet{Erk2010AFC} were, at the best of our knowledge, the first authors to measure the correlation between human-elicited thematic fit ratings and the scores assigned by a syntax-based Distributional Semantic Model (DSM). More specifically, their gold standard consisted of the human judgments collected by \citet{McRae1998ModelingTI} and \citet{pado2007integration}. The plausibility of each verb-filler pair was computed as the similarity between new candidate nouns and previously attested exemplars for each specific verb-role pairing (as already proposed in \citet{Erk2007ASS}). %Their distributional data came from two different corpora: they collected co-occurrence information from a 'primary' corpus, and then they used a 'generalization' corpus to compute similarities between argument fillers.

\citet{Baroni:2010:DMG:1945043.1945049} evaluated their Distributional Memory (henceforth DM)\footnote{In this paper, we will make reference to two different models of DM: DepDM and TypeDM. DepDM counts the frequency of dependency links between words (e.g. \textit{read, obj, book}), while TypeDM uses the variety of surface forms that express the link between words, rather than the link itself.} framework on the same datasets, adopting an approach to the task that has become dominant in the literature: for each verb role, they built a prototype vector by averaging the dependency-based vectors of its most typical fillers. The higher the similarity of a noun with a role prototype, the higher its plausibility as a filler for that role. \citet{Lenci_composingand} has later extended the model to account for the dynamic update of the expectations on an argument, depending on how %\esnote[margin,noinline]{"depending on how"?} 
another role is filled. By using the same DM tensor, this study tested an additive and a multiplicative model \citep{mitchell2010composition} to compose and update the expectations on the patient filler of the subject-verb-object triples of the Bicknell dataset \citep{bicknell2010effects}. %\esnote[margin,noinline]{Maybe we can add here that DM is not completely unsupervised, so that we can recall it later in the results? EC: boh, dici che la cosa sul fatto che DM usa regole handcrafted non basta? Cosa si può aggiungere?}

The thematic fit models proposed by \citet{Sayeed2014combining} and \citet{Sayeed2015AnEO} are similar to Baroni and Lenci's, but their DSMs were built by using the roles assigned by the SENNA semantic role labeler \citep{Collobert:2011:NLP:1953048.2078186} to define the feature space. These authors argued that the prototype-based method with dependencies works well when applied to the agent and to the patient role (which are almost always syntactically realized as subjects and objects), but that it might be problematic to apply it to different roles, such as instruments and locations, as the construction of the prototype would have to rely on prepositional complements as typical fillers, and the meaning of prepositions can be ambiguous. %\esnote[margin,noinline]{Qui e' d'obbligo mettere che noi dimostriamo il contrario in questo paper: è un punto narrativo fortissimo, che sarebbe da riprendere anche nelle conclusioni: abbozzo sotto e lo lascio a te} 
%Therefore, Sayeed and colleagues concluded that features based on semantic roles were more suitable for the task.
Comparing their results with \citet{Baroni:2010:DMG:1945043.1945049}, the authors showed that their system outperforms the syntax-based model DepDM and almost matches the scores of the best performing TypeDM, which uses hand-crafted rules. Moreover, they were the first to evaluate thematic role plausibility for roles other than agent and patient, as they computed the scores also for the instruments and for the locations of the Ferretti datasets \citep{Ferretti2001516}. %\textbf{In this paper, we will show that methods based on dependencies can also be applied with good results to roles other than agent and patient.}
%\ecnote[margin,noinline]{Tieni conto che queste sono comunque affermazioni MOLTO forti sulla performance del sistema. Poi, la storia della disambiguazione su quali basi potremmo dimostrare che il sistema funziona bene da questo punto di vista?}

%The TypeDM model and the role-based model were further developed by Greenberg et al. \shortcite{Greenberg:15a,Greenberg:15b}, who investigated 
\citet{Greenberg2015VerbPA,Greenberg2015ImprovingUV} further developed the TypeDM and the role-based models, investigating the effects of verb polysemy on human thematic fit judgments and introducing a hierarchical agglomerative clustering algorithm into the prototype creation process. %\esnote[margin,noinline]{Di nuovo, si potrebbe mettere il seme della capacita disambiguativa del nostro metodo} 
Their goal was to cluster together typical fillers into multiple prototypes, corresponding to different verb senses, and their results showed constant improvements of the performance of the DM-based model. %\textbf{Polysemy is dealt in our system through a combination of salience ranking and intersection for the vector features, which is likely to favor those features related to the most appropriate word sense in a given verb-argument pair (see Section 3.3).}

Finally, \citet{Tilk2016EventPM} presented two neural network architectures for generating probability distributions over selectional preferences for each thematic role. Their models took advantage of supervised training on two role-labeled corpora to optimize the distributional representation for thematic fit modeling, and managed to obtain significant improvements over the other systems on almost all the evaluation datasets. They also evaluated their model on the task of composing and updating verb argument expectations, obtaining a performance comparable to \citet{Lenci_composingand}.

%\ecnote[margin,noinline]{Non sono molto d'accordo nel mettere queste note nel related work, per me sarebbero molto fuori posto. Questa sezione poi è stata apprezzata da tutti i reviewer che hanno recensito il paper, non vedrei motivo di cambiarla più di tanto.}
%\esnote[margin,noinline]{Ok... Volevo creare l'aspettativa, ma sono d'accordo. Spostato sotto, in neretto}

\section{Methodology}

As pointed out by \citet{Sayeed2016ThematicFE}, most works on unsupervised thematic fit estimation vary in the method adopted for constructing the prototypes. The semantic role prototype is usually a vector, obtained by averaging the most typical fillers, and plausibility of new fillers depends on their similarity to the prototype, assessed by means of vector cosine (the standard similarity measure for DSMs; see \citet{Turney2010FromFT}). % \shortcite{Turney:10}).  

Its merits notwithstanding, we argue that this method is not optimal for characterizing roles. Distributional vectors are typically built as out-of-context representations, and they conflate different senses. By building the prototype as the centroid of a cluster of vectors and measuring then the thematic fit with vector cosine, the plausibility score is inevitably affected by many contexts that are irrelevant for the specific verb-argument combination.\footnote{For an overview on the limitations of vector cosine, see: \citet{Li2013,Dinu2014ImprovingZL,Schnabel2015EvaluationMF,Faruqui2016ProblemsWE,santus2016testing}.} %\esnote[margin,noinline]{Da aggiungere i reference se piacciono}
This is likely to be one of the main reasons behind the difficulties of modeling roles other than agent and patient with syntax-based DSMs. We claim that improving the prototype representation might lead to a better characterization of thematic roles, and to a better treatment of polysemy.

%\textbf{In this paper, we will show that methods based on dependencies can also be applied with good results to roles other than agent and patient.}
%\textbf{Polysemy is dealt in our system through a combination of salience ranking and intersection for the vector features, which is likely to favor those features related to the most appropriate word sense in a given verb-argument pair (see Section 3.3).}

When a verb and an argument are composed, humans are intuitively able to select only the part of the potential meaning of the words that is relevant for the concept being expressed (e.g. in \textit{The player hit the ball}, humans would certainly exclude from the meaning of \emph{ball} semantic dimensions that are strictly related to its dancing sense). In other words, not all the features of the semantic representations are active, and the composition process makes some features more `prominent', while moving others to the background.\footnote{An early proposal going in this direction is the predication theory by \citet{Kintsch2001Predication1P}, which exploited Latent Semantic Analysis to select only the vector features that are appropriate for predicate-argument composition. }%See also \citet{Erk:Pado:2008} for a syntax-based distributional model of word meaning in context.}

%\ecnote[margin,noinline]{Volevo aggiungere un'altra nota su un riferimento bibliografico interessante, ma lo commento per guadagnare spazio. Dopotutto, possiamo eventualmente reinserirlo in fase di camera-ready.}
%%\esnote[margin,noinline]{Chi? Hampton 2007? Non prende spazio}
%\esnote[margin,noinline]{This paragraph should be expanded; we should probably speak about "noise" of features, as they might be irrelevant; for what concerns the disambiguation, I believe that ranking by salience might take on top the features related to the most prototypical sense... Can we test it? We might see whether "runs" has something like "marathon, program, dog, pc" among the features}
%Intuitively, the composition between a verb and an argument makes some semantic features more prominent than others. At the best of our knowledge, such hypothesis has not been tested at the experimental level. However, we think that verb-argument composition could share the same process with the modifier-head constructions that are at the center of interest of the studies on conceptual combinations \cite{Hampton:07}. 
Although we are not aware of experimental works specifically dedicated to verb-argument composition, a similar idea has been supported in studies on conceptual combinations \citep{hampton1997conceptual,Hampton2007TypicalityGM}: when a head and a modifier are combined, their interaction affects the saliency of the features in the original concepts. For example, in \textit{racing car}, the most salient properties would be those related to SPEED, whereas in \textit{family car} SPACE properties would probably be more prominent.
\citet{Yeh2006TheSN} used a property priming experiment to show how the concept features activated during language comprehension vary across the background situations described by the sentence they occur in. When concepts are combined in a sentence, the features that are relevant for the specific combination are activated and are then easier to verify for human subjects. %In their work, experimental subjects were faster in recognizing \textit{can be walked upon} as a property of \textit{roof} in relevant situations \ref{exa} than in irrelevant situations \ref{exb}\textbf{, proving that situational context causes some properties of the conceptual representations to become more salient than others}.

%\ex.
%\a. \label{exa} \textit{The roof creaked under the weight of the repairman.}
%\b. \label{exb} \textit{The roof had been renovated prior to the rainy season.}

%In other words, the situational context causes some properties of the conceptual representations to become more salient than others.
%\esnote[margin,noinline]{Ho difficolta a capire come questi due paper siano supportive del paragrafo precedente. Puoi metterlo piu esplicito? Se invece vuoi usarli per argomentare la nostra seconda assunzione, sii piu esplicito anziché usare "interestingly"}
The same could be true for linguistically-derived properties of lexical meaning: \citet{Simmons2008fMRIEF} brought neuroimaging evidence of the early activation of word association areas during property generation tasks, and \citet{santos_chaigneau_simmons_barsalou_2011} showed that word associates are often among the properties generated for a given concept.
%\esnote[margin,noinline]{E' questa la frase in cui mancava il verbo? l'hai sistamata gia?}
%\ecnote[margin,noinline]{Yes}
Such findings suggest that, while we combine concepts, both embodied simulations and word distributions influence property salience \citep{barsalou2008language}.
%AL: questo paragrafo sopra non mi sembra rilevante per il tema

Our model makes the following assumptions:
\begin{itemize}
\item the composition between a verb role representation
%(whatever that might be)
and an argument shares the same cognitive mechanism underlying conceptual combinations;
%\footnote{The literature in the field has especially focused on modifier-head combinations \cite{Hampton:96,Hampton:07}. However, we are not aware of studies specifically dedicated to predicate-argument feature composition}
\item at least part of semantic representations is derived from, and/or mirrored in, linguistic data.\footnote{See also the so-called 'strong version' of the Distributional Hypothesis \citep{MillerAndCharles,lenci2008distributional}.}
%as suggested by the results of the above-mentioned studies on property generation\esnote[margin,noinline]{A quali ti riferisci esattamente? A quelli di Santos e di Simmons}\footnote{See also the proposal of Vigliocco et al. \shortcite{Vigliocco:09} for a general theory of meaning representation based on the joint contribution of sensory-motor and linguistic information.}.
Consistently, the process of selecting the relevant features of the concepts being composed corresponds to modify the salience of the dimensions of distributional vectors;
\item thematic fit computation is carried out on the basis of the activation and selection of salient features of a verb thematic role prototype and of the candidate argument filler vectors.
\end{itemize}

We rely on syntax-based DSMs, using dependency relations to approximate verb-specific roles and to identify their most typical fillers: for agents/patients, we extract the most frequent subjects/objects, for instruments we use the prepositional complements introduced by \textit{with}, and for locations those introduced by either \textit{on, at} or \textit{in}.

Assuming that the linguistic features of distributional vectors correspond to the properties of conceptual composition processes, a candidate filler can be represented as a sorted distributional vector of the filler term, in which the most salient contexts occupy the top positions. Similarly, the abstract representation of a verb-specific role is a sorted prototype-vector, whose features derive from the sum of the most typical filler vectors for that verb-specific role.

Differently from Baroni and Lenci, the core and novel aspect of our proposal, described in the following subsections, is that we do not simply measure the correlation between all the features of candidate and prototype vectors (as vector cosine would do on unsorted vectors), but rather we \textit{rank} and \textit{filter} the features, computing the \textit{weighted overlap} with a rank-based similarity measure inspired by $APSyn$, a recent proposal by \citet{santus2016testing,santus2016unsupervised,santus2016nerd} which has shown interesting results in synonymy detection and similarity estimation. As we will show in the next sections, the new metric assigns high scores to candidate fillers sharing many salient contexts with the verb-specific role prototype.

%and we apply a similarity measure inspired by \citet{santus2016unsupervised} allowing us to \textit{filter} and \textit{weight} the contexts according to their salience for a given thematic role, and to assign high scores to new candidate fillers sharing many salient contexts with the role prototype.

%\textbf{I HAVE COMMENTED THIS PART AS IT MIGHT BE SUFFICIENT MENTIONING THE FILTER ABOVE, AS PART OF THE ALGORITHM.} 

%PUOI ANCHE DIRE CHE ABBIAMO NOTATO CHE PER OGNI ROLE, UNA CERTA TIPOLOGIA DI FEATURES è PIU RILEVANTE... (PRED/COMPL, ETC.) E CHE QUINDI NEL NOSTRO METODO PREVEDIAMO ANCHE LA POSSIBILITà DI INTRODURRE UN FILTRO PER TIPOLOGIA

%\textit{a}) a method for ranking the contexts according to their salience for a specific verb role and of \textit{b}) a similarity measure assigning more weight to salient contexts shared by the prototype and by the candidate filler. 
%DEFINE THE TERM PROTOTYPE, OTHERWISE I'll GIVE YOU 1.5: LOL. 

%\footnote{The notion of \textit{salience} in linguistics and in cognitive science is a very ambiguous one. For the sake of clarity, by \textit{salience} we mean something close to Schmid and G{\"u}nther \shortcite{Schmid:16}'s definition of \textit{salience by contextual entrechment}, i.e. an object is salient when it matches the expectations triggered by the current context. \\ In our proposal, the 'salient' features for a given verb-specific slot correspond to distributional contexts that are strongly associated with a number of the fillers of the corresponding role, and are therefore expected to be associated also with new candidate fillers.} 

\begin{table*}[t]
\tiny
\centering
\resizebox{\textwidth}{!}{
\begin{tabular}{|p{20mm}|p{50mm}|p{40mm}|}
\hline
\multicolumn{1}{|c|}{} & \multicolumn{1}{c|}{\textbf{Typical Fillers}} & \multicolumn{1}{c|}{\textbf{Top Second Order Contexts}}  \\ \hline
\textbf{subject: cure-v} & treatment-n, drug-n, resin-n, doctor-n, surgery-n, medicine-n, therapy-n, antibiotic-n, dose-n, operation-n, water-n... & obj-1:prescribe-v, sbj-1:prescribe-v, sbj-1:prevent-v, sbj-1:contraindicate-v, [...] \\ \hline
\textbf{object: abandon-v} & plan-n, idea-n, project-n, attempt-n, position-n, principle-n, policy-n, ship-n, practice-n, hope-n, fort-n, claim-n... & obj-1:revive-v, obj-1:defend-v, obj-1:renounce-v, obj-1:espouse-v, sbj-1:entail-v... \\ \hline
\textbf{instrument: eat-v} & bread-n, hand-n, spoon-n, sauce-n, relish-n, fork-n, finger-n, meal-n, knife-n, friend-n, chopstick-n, rice-n, food-n... & obj-1:flavour-v, obj-1:taste-v, obj-1:spoon-v, sbj-1:taste-v, obj-1:slice-v in:bowl-n... \\ \hline
\textbf{location: walk-v} & in:direction-n, at:time, at:pace-n, on:path-n, at:night, on:side-n, at:end, on:beach-n, on:leg, in:area, in:way... & obj-1:wander-v, obj-1:stroll-v, obj-1:litter-v, obj-1:sweep-v, sbj-1:slope-v, obj-1:tread-v...\\
\hline
\end{tabular}}
\tiny\caption{Typical fillers and top \textit{second order contexts} for several verb-specific roles.}
\label{top_fillers_contexts}
\end{table*}

%\ecnote[margin,noinline]{Propongo di modificare la tabella così: prendere diversi verbi-slot, tipo treat-v\_obj, eat-v\_obj, e metterli sulle righe della tabella. Nella prima colonna ci mettiamo 2-3 typical fillers per ciascun verbo-slot, e nella seconda ci mettiamo un po' di contesti di secondo ordine. Inoltre, dai filler vanno tolte le funzioni sintattiche.}

\subsection{Typical Fillers}
\label{typical_fillers_sec}
The first step of our method consists in identifying the typical fillers of a verb-specific role. Following \citet{Baroni:2010:DMG:1945043.1945049}, we weighted the raw co-occurrences between verbs, syntactic relations and fillers in the TypeDM tensor of DM with Positive Local Mutual Information (PLMI; \citet{Evert2004TheSO}).

Given the co-occurrence count $O_{vrf}$ of the verb $v$, a syntactic relation $r$ and the filler $f$, we computed the expected count $E_{vrf}$ under the assumption of statistical independence:

\begin{equation} \small \label{1}
 PLMI(v,r,f) = log\left ( \frac{O_{v,r,f}}{E_{v,r,f}}\right ) * O_{v,r,f}
 \end{equation}

% \alnote{It is strange and odd to define PLMI as O/E ratio and then PPMI below with probability. LMI  is just PMI times O. I suggest to define them together.}

\noindent{}From the ranked list of \textit{(v,r,f)} tuples, for each slot, we selected as typical fillers the top $k$ lexemes with the highest PLMI scores (see examples in Table \ref{top_fillers_contexts}, \textit{Typical Fillers} column). In our experiments, we report results for $k=\{10, 30, 50\}$.

%Baroni and Lenci summed the typical filler TypeDM vectors for the corresponding syntactic slots. %, and they obtained their vectors from a dependency-based DSM enriched with manually selected lexico-syntactic patterns. 
%We built instead a purely dependency-based DSM (i.e. we only have features of the \textit{relation:word} type, such as \textit{sbj:dog,obj:apple etc.}) from which we extracted the filler vectors that we summed up to build the prototype, as well as those representing the fillers. 

\subsection{Role Prototype Vectors}

%Our method relies on a dependency-parsed corpus. We used syntactic relations to approximate verb-specific roles and to identify their most typical fillers: for agents/patients, we extracted the most frequent subjects/objects, for the instruments we used the prepositional complements introduced by \textit{with} and for the locations those introduced by either \textit{on, at} or \textit{in}.

To represent the typical fillers, the candidate fillers and the verb-specific role prototypes (which are obtained by summing their typical filler vectors), we built a syntax-based DSM. This includes \textit{relation:word} contexts, like \textit{sbj:dog, obj:apple, etc.}.

Contexts were weighted with Positive Pointwise Mutual Information (PPMI; \citet{Church:1990:WAN:89086.89095}, \citet{Bullinaria2012}, \citet{Levy2015ImprovingDS}). Given a context $c$ and a word $w$, the PPMI is defined as follows:

\begin{equation} \small \label{2}
PPMI(w,c) = max(PMI(w,c), 0)
\end{equation}

\begin{equation} \small \label{3}
PMI(w,c) = log \left (\frac{P(w,c)}{P(w)Ã—P(c)}\right ) = log\left ( \frac{|w,c|Ã—D}{|w|Ã—|c|}\right )
\end{equation}
 
\noindent{}where \textit{w} is the target word, \textit{c} is the given context, \textit{P(w,c)} is the probability of co-occurrence, and \textit{D} is the collection of observed word-context pairs.\footnote{A variant of this DSM weighted with PLMI (which is simply the PPMI multiplied by the word-context frequency)
was also built, but because of its lower and inconsistent performance we will not discuss it further. \citet{santus2016nerd} previously showed that their rank-based measure performs worse on PLMI-weighted vectors, as they are biased towards frequent contexts.} %It has been often used in the DSM literature because it smooths the bias of PPMI towards rare events. %It consists in multiplying the PPMI of \textit{w} and \textit{c} by their co-occurrence frequency.

The context $c$ of the prototype vector $P$ representing a thematic role has a value corresponding to the sum of the values of $c$ for each of the $k$ typical fillers used to build $P$. %, where $k$ corresponds to the number of fillers used to build such prototype. 
The contexts of $P$ are then sorted according to their weight. Desirably, the highest-ranking contexts for a role prototype will be those that are more strongly associated with many of its typical fillers. Such \textit{second order contexts} correspond to the most salient features of the verb-specific thematic role, as they are salient for many role fillers (some examples are reported in Table \ref{top_fillers_contexts}, \textit{Top Second Order Contexts} column).
%In order to ensure the relevance of these fillers in the prototype, we also filtered them by the most relevant syntactic relation for a given role (see Section \ref{filtering}).

In summary, we built centroid vectors for our verb-specific thematic roles by means of \textit{second order contexts}, which are first order dependency-based contexts of the most typical fillers of a verb-specific role. Since we are interested only in the most salient contexts, we ranked the centroid contexts according to their PPMI score,
%we filtered them (see Section \ref{filtering}), 
and we took the resulting rank as a distributional characterization of the thematic roles.

\subsection{Filtering the Contexts}
\label{filtering}

%\ecnote[margin,noinline]{Questa potrebbe essere la sezione giusta per inserire ipotesi riguardo alla selezione dei contesti. Nel caso, si potrebbe togliere l'ultima riga da tutte le celle della tabella 1 per guadagnare spazio.}
%\esnote[margin,noinline]{Rivedi il primo paragrafo: e' piu forte?}
%\ecnote[margin,noinline]{Non darei per scontato quel "it is reasonable that", la ragione per cui dovrebbero essere efficaci andrebbe chiarita meglio. Io limiterei l'esempio ai predicati, visto anche che nel paragrafo dopo ci riferiamo a quelli. Ora meglio,secondo me: sarebbe anche coerente con l'esempio del paragrafo dopo.}
%\esnote[margin,noinline]{Agreed}

Filtering the prototype dimensions according to syntactic criteria might be useful to improve our role representations.
%The contexts in the prototype vector can also be filtered according to syntactic criteria, in order to better represent the thematic role. 
It is, indeed, reasonable to hypothesize that predicates co-occurring with the typical patients of a verb are more relevant for the characterization of its patient role than -- let's say -- prepositional complements, as they correspond to other actions that are typically performed on the same patients.

Imagine that \textit{apple, pizza, cake} etc. are among the most salient fillers for the OBJ slot of \textit{to eat}, and that \textit{OBJ-1:slice-v, OBJ-1:devour-v, SBJ:kid-n, INSTRUMENT:fork-n, LOCATION:table-n} are some of the most salient contexts of the prototype.\footnote{Our DSM also makes use of inverse syntactic dependencies: \textit{target SYN-1 context} means that \textit{target} is linked to \textit{context} by the dependency relation \textit{SYN} (e.g. \textit{meal OBJ-1 devour} means that \textit{meal} is \textit{OBJ} of \textit{devour}).} Things that are
%We could decide that we want to assess the thematic fit of a new filler only on the basis of the predicates that it shares with the prototype, ignoring all the other syntactic relations. Consequently, 
typically \textit{sliced} and/or \textit{devoured} are more likely to be good fillers for the patient role \textit{to eat} than things that are simply located on a \textit{table} or that are patients of actions performed by \textit{kids}.
%be good fillers for the patient role of \textit{to eat}.
%\ecnote[margin,noinline]{Non ricordo, l'abbiamo già spiegato da qualche parte cosa sono le relazioni sintattiche inverse? In caso contrario, sarebbe meglio inserire una noticina in questo paragrafo, rimandando alla notazione di Distributional Memory.}
%In other words, we hypothesize that the prototype of the subject/object and the candidate filler should be compared in the predicates they share, rather than in any other context they might co-occur in. In the same fashion, we hypothesize that the prototype of instrument and location role and the eventual filler of such role are better characterized by the relevant prepositional contexts (e.g. with for instrument and at/on/.. for location).
%\esnote[margin,noinline]{Changed}
To test this hypothesis, we evaluated the performance of the system in three different settings, each of which selecting:
\begin{itemize}
\item only predicates in a subject/object relation (\textbf{SO} setting);
\item only prepositional complements (\textbf{PREP} setting);
\item both of them (\textbf{ALL} setting).
\end{itemize}

\subsection{Computing the Thematic Fit}

Our hypothesis is that fillers whose salience-ranked vector has a large overlap with the prototype representation should have a high thematic fit. Such overlap should take into account not only the number of shared features, but also their respective ranks in the salience-ranked vectors.
%\esnote[margin,noinline]{Secondo me questi due-tre paragrafi sono un punto delicato che dovrebbe essere spiegato meglio... Magari anche la formula andrebbe rivista appena...}
%\ecnote[margin,noinline]{Io farei giusto piccole modifiche di chiarimento alla fine di questo paragrafo}

When the prototype has been computed and the candidate filler vector has also been sorted, we can measure the \textit{Weighted Overlap} by adapting $APSyn$ \citep{santus2016testing,santus2016unsupervised,santus2016nerd} to our needs:

\begin{small}
\begin{equation} \label{apsyn}
%\resizebox{0.5\textwidth}{!}{
WO(w_{x}, w_{y}) =\sum_{\forall f\epsilon (x_{[1:N]}\cap y_{[1:N]})}\frac{1}{avg(r_{x}(f), r_{y}(f))}
\end{equation}
\end{small}

\noindent{}where for every feature $f$ in the intersection between the top $N$ features of the sorted vectors $x$, $x_{[1:N]}$, and $y$, $y_{[1:N]}$, we sum 1 divided by the average rank of the shared feature in $x$ and $y$, $r_{x}(f)$ and $r_{y}(f)$ ($N$ is a tunable parameter).

%\esnote[margin,noinline]{Please review}
%\ecnote[margin,noinline]{Looks fine to me. Maybe we can avoid repeating "the sorted vectors" two times.}
This measure assigns the maximum score to vectors sharing exactly the same dimensions, in the same salience ranking. %If the dimensions are shared but in different order (i.e. meaning that they have different salience for the two terms $x$ and $y$), the score is penalized.
The lower the rank of a shared context in the sorted vector, the smaller its contribution to the thematic fit score. If the feature set intersection is empty, the score will be 0.

Differently from cosine similarity, which conflates multiple senses, measuring the \textit{Weighted Overlap} between prototype and candidate filler can improve the estimation of the thematic fit by favoring the appropriate word senses: for example, for a verb-argument pair like \textit{embrace-v--communism-n}, \textit{communism-n} is likely to intersect and to increase the saliency (through the average rank) only of the second-order features of \textit{embrace-v} referring to its abstract sense.

\section{Experiments}

%\subsection{Thematic fit estimation}

\textbf{Datasets}. We tested our method on three popular datasets for thematic fit estimation, namely \citet{McRae1998ModelingTI}, \citet{Ferretti2001516} and \citet{pado2007integration}. All the datasets contain human plausibility judgments for verb-role-filler triples. McRae and Pad\'o include scores for agent and patient roles, whereas Ferretti includes instruments and locations (see Table \ref{coverage} for the coverage of each system for the datasets).

\noindent{}\textbf{Metrics}. Performance is evaluated as the Spearman correlation between the scores of the systems and the human plausibility judgments.

\begin{table}
\tiny
\centering
\resizebox{\columnwidth}{!}{
\begin{tabular}{|c|c|c|c|c|c|}
\hline
\textbf{Data} & \textbf{Our system} & \textbf{BL2010} & \textbf{SD2014} & \textbf{G2015} & \textbf{T2016} \\
\hline
Pad\'o & 96 & 100 & 99 & 100 & 99 \\ 
\hline
McRae & 100 & 95 & 96 & 95 & 96\\
\hline
Instr. & 100 & 93 & 94 & 93 & 94\\
\hline
Loc. & 96 & 99 & 100 & 99 & 100\\
\hline
\end{tabular}}
\tiny\caption{Dataset coverage (\%) for all systems.}
\label{coverage}
\end{table}

\begin{table*}[ht]
%\tiny
\centering
\resizebox{\textwidth}{!}{
\begin{tabular}{|c|c|c|c|c|c|c|c|c|c|c|c|c|c|c|}
\hline
\multirow{2}{*}{\textbf{Weight}} & \multirow{2}{*}{\textbf{N}} & \multirow{2}{*}{\textbf{\# Fillers}} & \multicolumn{3}{c|}{\textbf{Pad\'o}} & \multicolumn{3}{c|}{\textbf{Mcrae}} & \multicolumn{3}{c|}{\textbf{Ferretti - Instruments}} & \multicolumn{3}{c|}{\textbf{Ferretti - Locations}} \\ \cline{4-15} 
 &  &  & \textbf{ALL} & \textbf{SO} & \textbf{PREP} & \textbf{ALL} & \textbf{SO} & \textbf{PREP} & \textbf{ALL} & \textbf{SO} & \textbf{PREP} & \textbf{ALL} & \textbf{SO} & \textbf{PREP} \\ \hline
\multirow{3}{*}{\textbf{PPMI}} & \multirow{3}{*}{\textbf{2000}} & \textbf{10} & 0.43 & 0.45 & 0.26 & 0.25 & 0.27 & 0.19 & 0.43 & 0.41 & 0.46 & 0.25 & 0.27 & 0.28 \\ \cline{3-15} 
 &  & \textbf{30} & 0.47 & 0.49 & 0.33 & 0.26 & 0.28 & 0.22 & 0.42 & 0.41 & \textbf{0.50} & 0.28 & 0.31 & 0.37 \\ \cline{3-15} 
 &  & \textbf{50} & 0.46 & \textbf{0.50} & 0.35 & 0.27 & \textbf{0.29} & 0.24 & 0.39 & 0.38 & 0.47 & 0.28 & 0.32 & \textbf{0.39} \\ \hline
\multicolumn{2}{|c|}{\multirow{3}{*}{\textbf{\begin{tabular}[c]{@{}c@{}}Vector Cosine\\ (Baseline)\end{tabular}}}} & \textbf{10} & \multicolumn{3}{c|}{0.43} & \multicolumn{3}{c|}{0.25} & \multicolumn{3}{c|}{0.42} & \multicolumn{3}{c|}{0.29} \\ \cline{3-15} 
\multicolumn{2}{|c|}{} & \textbf{30} & \multicolumn{3}{c|}{0.47} & \multicolumn{3}{c|}{0.26} & \multicolumn{3}{c|}{0.41} & \multicolumn{3}{c|}{0.32} \\ \cline{3-15} 
\multicolumn{2}{|c|}{} & \textbf{50} & \multicolumn{3}{c|}{0.48} & \multicolumn{3}{c|}{0.26} & \multicolumn{3}{c|}{0.38} & \multicolumn{3}{c|}{0.31} \\ \hline
\multicolumn{15}{|c|}{\textbf{State of the Art}} \\ \hline
\multicolumn{3}{|c|}{\textbf{Baroni and Lenci (2010)}} & \multicolumn{3}{c|}{0.53} & \multicolumn{3}{c|}{0.33} & \multicolumn{3}{c|}{0.36} & \multicolumn{3}{c|}{0.23} \\ \hline
\multicolumn{3}{|c|}{\textbf{Sayeed and Demberg (2014)}} & \multicolumn{3}{c|}{0.56} & \multicolumn{3}{c|}{0.27} & \multicolumn{3}{c|}{0.28} & \multicolumn{3}{c|}{0.13} \\ \hline
\multicolumn{3}{|c|}{\textbf{Greenberg et al. (2015)}} & \multicolumn{3}{c|}{0.53} & \multicolumn{3}{c|}{0.36} & \multicolumn{3}{c|}{0.42} & \multicolumn{3}{c|}{0.29} \\ \hline
\multicolumn{3}{|c|}{\textbf{Tilk et al. (2016)}} & \multicolumn{3}{c|}{0.52} & \multicolumn{3}{c|}{0.38} & \multicolumn{3}{c|}{0.45} & \multicolumn{3}{c|}{0.44} \\ \hline
\end{tabular}}
\tiny\caption{Results for Pad\'o, McRae and Ferretti, Instruments and Locations, with $WO$ computed on PPMI matrix, varying the number of fillers (i.e. 10, 30 and 50) and the types of dependency contexts (i.e. ALL, SO and PREP). The best results of our system are in bold. A baseline reimplementing Baroni and Lenci (2010) -- with 10, 30 and 50 fillers -- and state of the art results from previous literature are reported for comparison.}
\label{results}
\end{table*}

\noindent{}\textbf{Fillers}. In order to make our results more comparable with previous studies, the typical fillers for each verb role were extracted from the TypeDM tensor of the Distributional Memory framework (see Section \ref{typical_fillers_sec}).\footnote{\url{http://clic.cimec.unitn.it/dm/}} Those were the same fillers used by \citet{Baroni:2010:DMG:1945043.1945049} and \citet{Greenberg2015ImprovingUV}.

\noindent{}\textbf{DSM}. Distributional information is derived from the concatenation of two corpora: the British National Corpus \citep{leech1992100} and Ukwac \citep{Baroni2009}.  Both were parsed with the Maltparser \citep{Nivre05maltparser:a}. From this concatenation, we built a dependency-based DSMs, weighted with PPMI, containing 20,145 targets (i.e. nouns and verbs with frequency above 1000) and 94,860 contexts.
%\footnote{An identical DSM weighted with PLMI was also built, but -- as discussed in Section \ref{result_discussion} -- it performed poorly and inconsistently, probably because PLMI is too biased towards the most frequent second order contexts.}
The syntactic relations taken into account were: \textit{sbj, sbj-1, obj, obj-1, at-1, in-1, on-1, with-1}.

\noindent{}\textbf{Settings}. To prove our hypotheses and verify the consistency of the system, we tested a large range of settings, varying:
\begin{itemize}
\item the number of fillers used to build the prototype, with the most typical values in the literature ranging between 10 and 50. We report the results for 10, 30 and 50 fillers%\ecnote[margin,noinline]{Metterei semplicemente che riportiamo i risultati per 10, 30 e 50 filler, senza scrivere altro.};
\item the types of the dependency relations used for calculating the overlap: we report results for the SO, PREP and ALL settings;
%\item the weighting measure used to rank the second order contexts: we tested both PPMI or PLMI, but we report results only for the former, as they are more stable and predictable, while briefly discussing also those obtained with PLMI.
\item the value of $N$, that is the number of top contexts that we take into account when computing the weighted overlap. Table \ref{results} reports the scores for our best setting, while the performances for other values of $N$ are discussed in the Section \ref{result_discussion}.
%\ecnote[margin,noinline]{Eventualmente, ricordati di aggiungere qui 2000, se abbiamo scores buoni}%\esnote[margin,noinline]{When we filter the contexts, we rarely have more than 1500 with the same syntactic relation; also, sometimes one of the two vectors has less than $N$, but the other still N; the current algorithm will still consider the intersection of the few into the top $N$ of the second}
\end{itemize}
\textbf{Baseline and State of the Art}. As a baseline, we use the thematic fit model by \citet{Baroni:2010:DMG:1945043.1945049}, with no ranking of the features of the prototypes and with vector cosine as a similarity metric.\footnote{This baseline is equivalent to the approach of \citet{Baroni:2010:DMG:1945043.1945049}, except for the fact that it is applied on a standard dependency-based DSM and not on TypeDM, which combines dependency links and handcrafted lexico-syntactic patterns: see Section \ref{related_work}.} Results are reported for 10, 30 and 50 fillers. %\ecnote[margin,noinline]{Specificare qui che si tratta di reimplementazione di Baroni e Lenci.}\esnote[margin,noinline]{Il fatto che non ci sia weighting nella baseline lo dobbiamo dichiarare? GUARDA LA NOTA A PIE PAGINA.} \ecnote[margin,noinline]{Per me no, diciamo che reimplementa metodo del coseno di Baroni-Lenci.} 
For reference, we also report the results of state-of-the-art models, both the unsupervised \citep{Baroni:2010:DMG:1945043.1945049,Sayeed2014combining,Greenberg2015ImprovingUV} and the supervised ones \citep{Tilk2016EventPM}. %For comparison, Table \ref{coverage} summarizes dataset coverage of each system.

\begin{figure*}
	\centering
  \includegraphics[width=14.5cm,height=9.5cm]{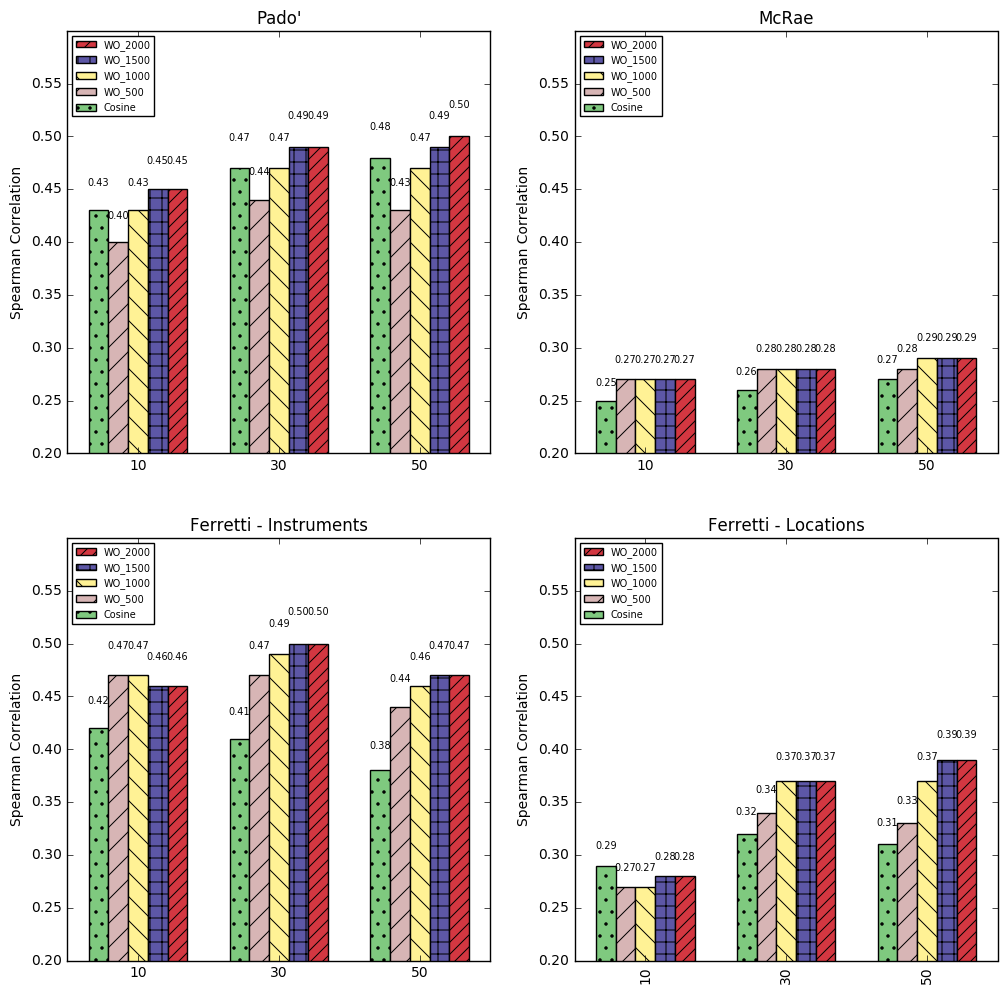}
  \tiny\caption{Results for Pad\'o, McRae and Ferretti, Instruments and Locations, with $WO$ (respectively SO and PREP) computed on PPMI matrix, varying the number of fillers (i.e. 10, 30 and 50) and the value of $N$ (i.e. 500, 1000, 1500 and 2000). A baseline reimplementing Baroni and Lenci (2010) -- with 10, 30 and 50 fillers -- is also reported in every test for comparison.}
  \label{performance_plot}
\end{figure*}

\begin{figure*}
	\centering
  \includegraphics[width=14.5cm,height=9.5cm]{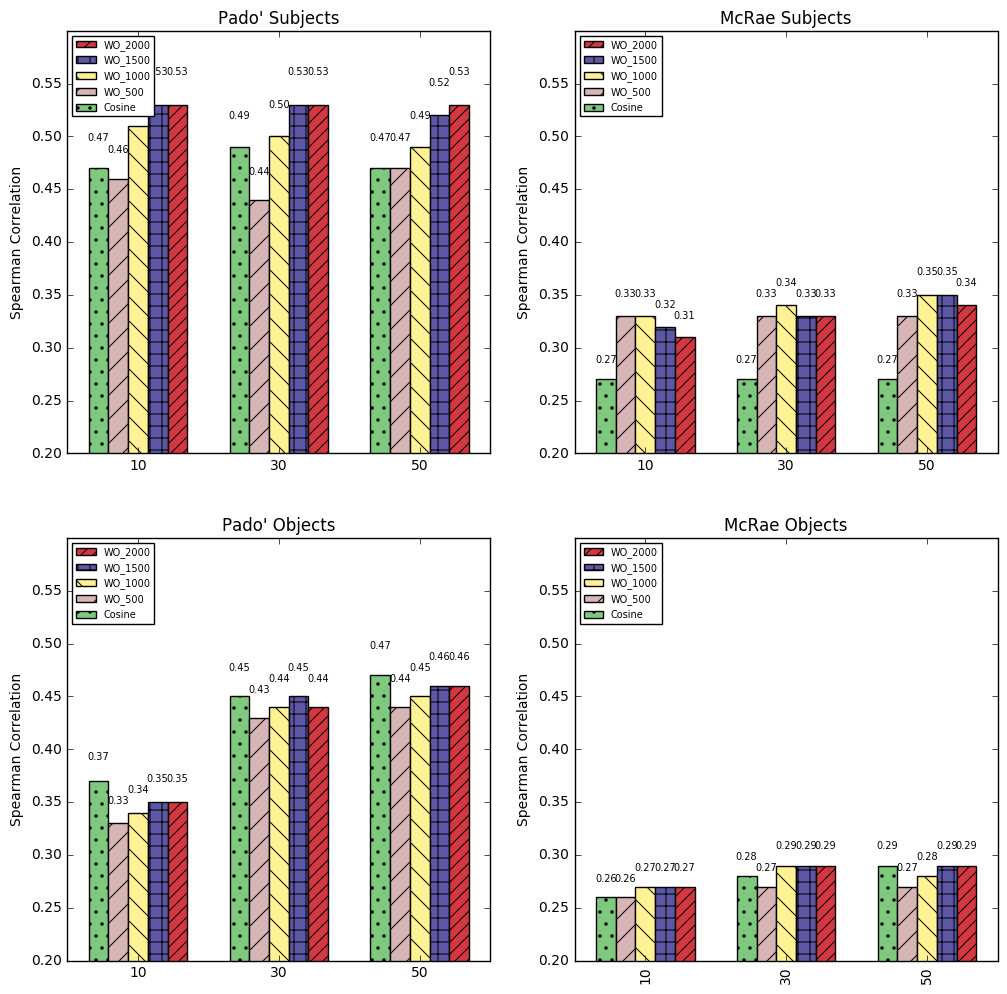}
  \tiny\caption{Results for the agent and patient roles in Pad\'o and McRae, with $WO$ (SO) computed on PPMI matrix, varying the number of fillers (i.e. 10, 30 and 50) and the value of $N$ (i.e. 500, 1000, 1500 and 2000). A baseline reimplementing Baroni and Lenci (2010) -- with 10, 30 and 50 fillers -- is also reported in every test for comparison.}
  \label{obj_sbj_plot}
\end{figure*}

%\esnote[margin,noinline]{Magari migliorare le figure}

\begin{table*}[ht]
%\tiny
\centering
\resizebox{\textwidth}{!}{
\begin{tabular}{cc|c|c|c|c|c|c|c|c|}
\cline{3-10}
\multicolumn{1}{l}{\multirow{2}{*}{}}                                                                                 & \textbf{}       & \multicolumn{4}{c|}{\textbf{BEST 35}}                                                  & \multicolumn{4}{c|}{\textbf{WORST 35}}                                               \\ \cline{2-10} 
\multicolumn{1}{l|}{}                                                                                                  & \textbf{Metric} & \textbf{Avg. Gold} & \textbf{Overlap} & \textbf{Syntax} & \textbf{Lexemes}                      & \textbf{Avg. Gold} & \textbf{Overlap} & \textbf{Syntax} & \textbf{Lexemes}                    \\ \cline{1-10} 
\multicolumn{1}{|c|}{\multirow{2}{*}{\begin{tabular}[c]{@{}c@{}}McRae\\ (k=50, Pred)\end{tabular}}}                   & Cos     & 4.90               &  3       & 14 sbj, 21 obj  & 3 sentence, 2 devour, 2 scratch...        & 4.75               &  26       & 24 sbj, 11 obj  & 2 consider, 2 entertain, 2 scrub           \\ \cline{2-10} 
\multicolumn{1}{|c|}{}                                                                                                & $WO$ 2000         & 4.20               &     4    & 17 sbj, 18 obj  & 2 haunt & 4.15               &    26     & 23 sbj, 12 obj  & 2 admire, 2 arrest, 2 consider, 2 entertain \\ \cline{1-10} 
\multicolumn{1}{|c|}{\multirow{2}{*}{\begin{tabular}[c]{@{}c@{}}Pad\'o\\ (k=50, Pred)\end{tabular}}}                  & Cos             & 4.07               &  10       & 12 sbj, 23 obj  & 4 advise, 4 eat, 4 embarrass          & 4.77               &    16     & 17 sbj, 18 obj  & 9 tell, 7 kill, 4 see               \\ \cline{2-10} 
\multicolumn{1}{|c|}{}                                                                                                & $WO$ 2000         & 4.35               &     10    & 21 sbj, 14 obj  & 3 confuse, 3 hear, 3 promise, 3 raise & 4.68               &    16     & 15 sbj, 20 obj  & 7 resent, 5 increase, 4 hear, 4 see \\ \cline{1-10} 
\multicolumn{1}{|c|}{\multirow{2}{*}{\begin{tabular}[c]{@{}c@{}}Ferretti - Instruments\\ (k=30, Compl)\end{tabular}}} & Cos             &     4.53               &   16      &         35 with        &   3 hung, 3 eat, 3 teach             &          4.51          &    22     &         35 with        &       4 repair, 3 teach, 3 inflate              \\ \cline{2-10} 
\multicolumn{1}{|c|}{}                                                                                                & $WO$ 2000         &          5.06          &      15   &         35 with        &         3 dig, 3 hunt          &         4.49           &    22     &    35 with             &       3 repair, 2 attract, 2 dig, 2 draw, 2 drink...                              \\ \cline{1-10} 
\multicolumn{1}{|c|}{\multirow{2}{*}{\begin{tabular}[c]{@{}c@{}}Ferretti - Locations\\ (k=50, Compl)\end{tabular}}}   & Cos             &       5.15             &     11    &         35 on/at/in        &           3 draw, 3 rescue      &        4.72            &    23     &        35 on/at/in         &      3 run, 2 wait, 2 wash, 2 shower...   \\ \cline{2-10} 
\multicolumn{1}{|c|}{}                                                                                                & $WO$ 2000         &         4.97           &    11     &        35 on/at/in         &        3 browse, 3 eat, 3 mingle, 3 rescue      &         4.47           &   23      &       35 on/at/in          &     3 run, 2 draw, 2 exercise, 2 shower, 2 wait...      \\ \cline{1-10} 
\end{tabular}}
\tiny\caption{Average gold values, number of items listed for both metrics, and distribution of syntactic and lexical forms among the 35 best and worst correlated items for every measure in the given datasets.}
\label{analysis}
\end{table*}

\section{Results}
\label{result_discussion}
Table \ref{results} describes the performance of the best setting (weight: PPMI; N=2000). In the first three rows, the table shows the scores obtained by our system varying the types of dependency contexts %used while calculating the $WO$ 
(i.e. ALL, SO, PREP) and the number of fillers considered for the prototype (i.e. 10, 30 and 50). The other rows respectively show i) the scores obtained by calculating the vector cosine between the role prototype vector (i.e. the vector obtained by summing the most typical fillers, with no salience ranking of the dimensions) and the candidate filler vector %, as it is in our DSM 
and ii) the scores reported in the literature for the best unsupervised and supervised models. %Vector cosine baseline is calculated for 10, 30 and 50 fillers to make comparison easier.

At a glance, our best scores always outperform the reimplementation of Baroni and Lenci, being mostly competitive with the state of the art models. More precisely, for agents and patients the performance is close to the reported scores for DM, when only predicates are used in the $WO$ calculation, as hypothesized in Section \ref{filtering}. The neural network of Tilk and colleagues retains a significant advantage on our models only for the McRae dataset. Our system, however, shows a remarkable improvements on the Ferretti's datasets, and specifically on Ferretti-Instruments, when only complements are used (see Section \ref{filtering}), outperforming even the supervised and more complex model by \citet{Tilk2016EventPM}, which has access to semantic roles information. 
%\ecnote[margin,noinline]{Per me bene cosi'. Anzi, mi piace piu' il claim qui che quelli in mezzo alla sezione del related work}
Compared to the other unsupervised models, our system has a statistically significant advantage over \citet{Baroni:2010:DMG:1945043.1945049} on the locations dataset and over \citet{Sayeed2014combining} on the locations and on the instruments dataset ($p < 0.05$).\footnote{p-values computed with Fisher's r-to-z transformation.} %\ecnote[margin,noinline]{Fare molta attenzione ai claims nel riformulare questa sezione}
%\esnote[margin,noinline]{Ti ricordi se avevi testato se i nostri risultati erano significativi rispetto alla baseline? Sarebbe un claim importante}
%\ecnote[margin,noinline]{Purtroppo non lo erano, anche se AP funziona costantemente meglio}

At the best of our knowledge, the result for the instruments is the best reported until now in the literature. This is particularly interesting because -- as pointed out by \citet{Sayeed2014combining} -- instruments and locations are difficult to model for a dependency-based system, given the ambiguity of prepositional phrases (e.g. \textit{with} does not only encode instruments, but it can also encode other roles, such as in \textit{I ate a pizza with Mark}). We think this is the main reason behind the different trend observed for the Instruments datasets with respect to the number of the fillers (see Table \ref{results} and Figure \ref{performance_plot}). Unlike all the other datasets, instrument prototypes built with more fillers tend to be more noisy and therefore to pull down both the vector cosine and $WO$ performance (this is partially true also for locations, where the performances -- for cosine and $WO$ with a lower number of contexts -- drop with more than 30 fillers: see Figure \ref{performance_plot}). Systems based on semantic role labeling have an advantage in this sense, as they do not have to deal with prepositional ambiguity.
%\ecnote[margin,noinline]{Controlla anche tu se sia vero quello che aggiunto, ovvero che il drop riguardi anche $WO$ con relativamente pochi contesti}

Our results show that, by weighting and filtering the features of the role prototype, dependency-based approaches can be successful in modeling roles other than agent and patient, eventually dealing also with the ambiguity of prepositional phrases.\\

\noindent{}\textbf{Settings.} Apart from the above-mentioned exceptions, the best scores are obtained building the prototypes with a higher number of fillers, typically with 50,
%(with the only exception of Ferretti-Instrument, where the prototype built with 30 fillers obtains a slightly better result)
and calculating the $WO$ only with a syntactically-filtered set of contexts. More specifically, Pad\'o and McRae benefit from the calculation of $WO$ using only second order subject-object predicates (i.e. SO), while Ferretti-Instruments and Ferretti-Locations benefit from the exclusive use of prepositional complements (i.e. PREP). On the other hand, the opposite setting (e.g. SO for Ferretti-Instruments and Ferretti-Locations and PREP for Pad\'o and McRae) leads to much lower scores, whereas the full vectors (i.e. ALL) tend to have a stable-but-not-excellent performances on all datasets.

As briefly mentioned above, in our experiments, we tested both PPMI and PLMI as weighting measures. Table \ref{results} only reports PPMI scores because it performs more regularly than PLMI, whose behaviour is often unpredictable.
%In fact, while we were able to clearly identify the contribution of the parameters with the PPMI-based matrix, the PLMI-based matrix produced unpredictable results while varying the parameters. 

%This might depend on the fact that PLMI is more sensitive than PPMI to frequent events, negatively affecting the salience ranking of the \textit{second order contexts} (as already noticed by \citet{santus2016nerd}). We leave a more in-depth analysis to a future and more extended research work.
%\alnote{This was already mentioned above and it is not important to repeat it again}

%\footnote{Some of the PLMI-based results were few points higher than those we report for PPMI. However, given the difficulty of establishing the relation between the parameters and the performance, we postpone such analysis to further studies.}%\esnote[margin,noinline]{IMPORTANTE: Questa nota e un po' delicata... Fatemi sapere}

A parameter that has an impact on the performance of our system is the value of $N$, which is the number of \textit{second order contexts} that are considered when calculating the $WO$. We have noticed that the performance of $WO$ is directly related to the growth of $N$, and this can be noticed in Figure \ref{performance_plot}, where $WO$ is plotted for the different values of $N$ with every combination of dataset and number of fillers. For space reasons, the plot only contains the performance for the best type of \textit{second order contexts} for each dataset (i.e. SO for Pad\'o and McRae and COMP for Ferretti-Locations and Ferretti-Instruments). As it can be seen in Figure \ref{performance_plot}, the scores of $WO$ tend to grow with the growth of $N$ in all datasets. Interestingly, they are largely above the competitive baseline in most of the cases, the only exceptions being Pad\'o (where a large $N$ is necessary to outperform the baseline) and Ferretti-Locations with 10 fillers (prepositional ambiguity might have caused the introduction of noisy fillers among the top ones).\\

\noindent{}\textbf{Agent \& Patient.} In order to further evaluate our system, we have split Pad\'o and McRae datasets into agent and patient subsets. 
%, respectively named \textit{sbj\_pado}, \textit{obj\_pado}, \textit{sbj\_mcrae} and \textit{obj\_mcrae}. 
Figure \ref{obj_sbj_plot} describes the performance of $WO$ and vector cosine baseline while varying $N$ and the number of fillers. The plot shows a clearly better performance of $WO$ for the agent role (i.e. \textit{subject}), especially when $N$ is equal or over 1000 (note that the value of $N$ has little impact in the agent subset of the McRae dataset).
%\alnote{From Fig. 2, it actually seems that the patient /obj performs better than the agent/subj, at least for the Pado. For the McRae is the reverse. Am I missing something?}
Such advantage, however, is reduced for the patient role (i.e. \textit{object}). This is particularly interesting because we do not observe large drops in performance for the vector cosine between agent and patient role (except for Pad\'o, $k=10$). The drop is particularly noticeable in Pad\'o, a dataset which has several non-constraining verbs (especially for the patient role: a similar observation was also made by \citet{Tilk2016EventPM}). As the constraints on the typical fillers of such verbs are very loose, we hypothesize that it is more difficult to find a set of salient features that are shared by many typical fillers. Therefore, estimations based on the whole vectors turn out to be more reliable. This can be confirmed by looking at the worst correlated words reported in \textit{Lexemes} column, in Table \ref{analysis}.
%\esnote[margin,noinline]{Qui si puo mettere di piu su l'omogeneita e la similarita dei fillers (obj/sbj)}
%\esnote[margin,noinline]{vedi se cosi va bene}

%\esnote[margin,noinline]{Serve un error analysis}
\subsection{Error Analysis}
We performed an error analysis to verify -- for the best settings of $WO$ in each dataset -- the correlation between vector cosine and $WO$ scores (see Table \ref{WO_VC_correlation}), and the peculiarities of the entries with the strongest and the weakest correlation (see Table \ref{analysis}).

We found that $WO$ and vector cosine always have a high correlation (i.e. above 0.80), with the highest correlations reported for McRae and Ferretti-Instruments. Looking at Table \ref{analysis} we can also observe that:
\begin{itemize}
\item the average gold value of the 35 most (4.65) and least (4.56) correlated items does not substantially differ from the average gold value calculated on the full datasets (4.31), meaning that the distribution of likely and unlikely fillers among the best and worst correlated items is similar to the one in the datasets (i.e. no bias can be identified);
\item both measures have difficulties on the same test items (probably because of loose semantic constraints), but report their best performances on different pairs (see \textit{Overlap} and \textit{Lexemes} columns);
\item syntactically, vector cosine correlates better with objects, while $WO$ is more balanced between objects and subjects, often showing a preference for the latter (see the distribution in \textit{Syntax} column).
\end{itemize}

\begin{table}
\small
\centering
\begin{tabular}{|c|c|}
\hline
\textbf{Dataset}     & \textbf{Correlation} \\ \hline
McRae                & 0.88                 \\ \hline
Pad\'o                 & 0.81                 \\ \hline
Ferretti - Instruments & 0.90                 \\ \hline
Ferretti - Locations  & 0.83                 \\ \hline
\end{tabular}
\tiny\caption{Correlation between $WO$ and vector cosine in $WO$ best settings for all datasets}
\label{WO_VC_correlation}
\end{table}

% TO ADD:
% - TABLE Subj/Obj and discussion of N/Fillers

%\ecnote[margin,noinline]{IMPORTANTE: quando si parla dei risultati, ricordiamoci di sottolineare la performance dei modelli in rapporto alla loro differenza di complessità e al tipo di input su cui lavorano, e alla velocità di esecuzione.}
%
\section{Conclusions}
In this paper, we have introduced an unsupervised distributional method for modeling predicate-argument thematic fit judgments which works purely on syntactic information. %We use a dependency-based DSM to build a prototypical representation of verb-specific roles and estimate how likely a candidate filler fits such role. 

The method, inspired by cognitive and psycholinguistic findings, consists in: i) extracting and filtering the most salient \textit{second order contexts} for each verb-specific role, i.e. the most salient semantic dimensions of \textit{typical verb-specific role fillers}; and then ii) estimating the thematic fit as a \textit{weighted overlap} between the top features of the candidate fillers and of the prototypes. Once tested on some popular datasets of thematic fit judgments, %(i.e. Pad\'o, McRae, Ferretti-Instruments and Ferretti-Locations), 
our method consistently outperforms a baseline re-implementing the thematic fit model of \citet{Baroni:2010:DMG:1945043.1945049} and proves to be competitive with state of the art models. It even registered the best performance on the Ferretti-Instruments dataset and it is the second best on the Ferretti-Locations, which were known to be particularly hard to model for dependency-based approaches.%\esnote[margin,noinline]{Forse dobbiamo migliorare un po' le conclusioni, discutendo maggiormente: N (unsupervised and tuning sono in contrasto), l'omogeneita e la similarita dei fillers (obj/sbj), la capacita di rappresentare role ambigui (hindrance), qualche conclusione cognitiva sulla rappresentazione?}

Our method is simple, economic and efficient, it works purely on syntactic dependencies (so it does not require a role-labeled corpus) and achieves good results even with no supervised training. Finally, it offers linguistically and cognitively grounded insights on the process of prototype creation and contextual feature salience, preparing the ground for further speculations and optimizations. For example, future work might aim at identifying strategies for tuning the parameter $N$ to account for the different degrees of selectivity of each verb-specific role. Another possible extension would be the inclusion of a mechanism for updating the role prototypes depending on how the other roles are filled, which would be the key for a more realistic and dynamic model of thematic fit expectations \citep{Lenci_composingand}.

\section*{Acknowledgments}

We would like to thank the anonymous reviewers for their helpful suggestions.

This work has been carried out thanks to the support of the A*MIDEX grant (n°ANR-11-IDEX-0001-02) funded by the French Government ``Investissements d'Avenir" program.
%The acknowledgments should go immediately before the references.  Do
%not number the acknowledgments section. Do not include this section
%when submitting your paper for review.

\bibliographystyle{emnlp_natbib}
\bibliography{emnlp2017}

\end{document}